\documentclass[sigcon]{acmart}
\AtBeginDocument{%
  }

\setcopyright{acmlicensed}
\copyrightyear{2018}
\acmYear{2018}
\acmDOI{XXXXXXX.XXXXXXX}
\acmConference[Conference acronym 'XX]{Make sure to enter the correct
  conference title from your rights confirmation email}{June 03--05,
  2018}{Woodstock, NY}
\acmISBN{978-1-4503-XXXX-X/2018/06}

\usepackage{listings}
\usepackage{xcolor}
\definecolor{promptbg}{RGB}{245,245,245}

\begin{document}


\title{Self-Filtered Distillation with LLM-generated \\Trust Indicators for Reliable Patent Classification}

\author{Yongmin Yoo}
\authornote{Equal contribution.}
\email{yooyongmin91@gmail.com}
\affiliation{%
  \institution{Frontier AI Research Centre, School of Computing, Faculty of Science and Engineering, Macquarie University}
  \city{Sydney}
  \state{NSW}
  \country{Australia}
}

\author{Xu Zhang}
\authornotemark[1]
\email{xu.zhang12@hdr.mq.edu.au}
\affiliation{%
  \institution{Frontier AI Research Centre, School of Computing, Faculty of Science and Engineering, Macquarie University}
  \city{Sydney}
  \state{NSW}
  \country{Australia}
}

\author{Longbing Cao}
\email{longbing.cao@mq.edu.au}
\affiliation{%
  \institution{Frontier AI Research Centre, School of Computing, Faculty of Science and Engineering, Macquarie University}
  \city{Sydney}
  \state{NSW}
  \country{Australia}
}
\renewcommand{\shortauthors}{Trovato et al.}

\begin{abstract}
Organizing large-scale patent corpora according to classification schemes is a core information management task that determines the accuracy and efficiency of prior art retrieval, technology knowledge discovery, and intellectual property decision-making. Recent approaches distill natural language rationales generated by large language models (LLMs) into compact student models, yet logical errors, label mismatches, and taxonomy misalignments inherent in these rationales are indiscriminately absorbed during training, undermining classification reliability and propagating errors throughout downstream information processes. Rather than correcting such errors post-hoc, we propose Self-Filtered Distillation (SFD), which embeds quality assurance directly into the learning process by reinterpreting LLM-generated rationales as trust indicators rather than ground-truth supervision. SFD integrates three unsupervised signals into a unified trust score that dynamically modulates each training instance's contribution: Self-Consistency, which quantifies agreement among independently generated rationales; Class Entailment Alignment, which evaluates semantic coherence between a rationale and its assigned CPC class definition; and LLM Agreement Scoring, which assesses external plausibility through an independent verifier. On the USPTO-2M benchmark comprising over two million patents, SFD achieves up to 38.7\% relative improvement in Macro-F1 across four student architectures, and the strong correlation between trust scores and expert judgments ($r = 0.685$) confirms that the framework provides not only accurate predictions but also decomposable confidence semantics that enable auditable and self-documenting classification outcomes for large-scale patent knowledge organization.
\end{abstract}


\maketitle

\section{Introduction}
With over 4.5 million patent applications filed worldwide each year, accurately organizing these documents into technical classification schemes is essential for prior art retrieval, technology landscape analysis, and intellectual property decision-making~\citep{jiang2025natural,yoo2025patentmind}. Yet patents pose unique challenges for automated classification: highly specialized terminology, complex claim structures, and strategically ambiguous expressions make them fundamentally harder to categorize than general-domain texts. Recent advances in large language models (LLMs) offer strong potential to address these challenges by processing massive patent datasets with improved accuracy~\citep{lee2020patent,bai2024patentgpt}. Moreover, LLMs can generate natural language rationales that enhance interpretability and provide a reasoning-based foundation for trust in classification outcomes~\citep{wei2022chain,hsieh2023distilling}.

\begin{figure}[t]
  \centering  \includegraphics[width=0.99\linewidth,keepaspectratio]{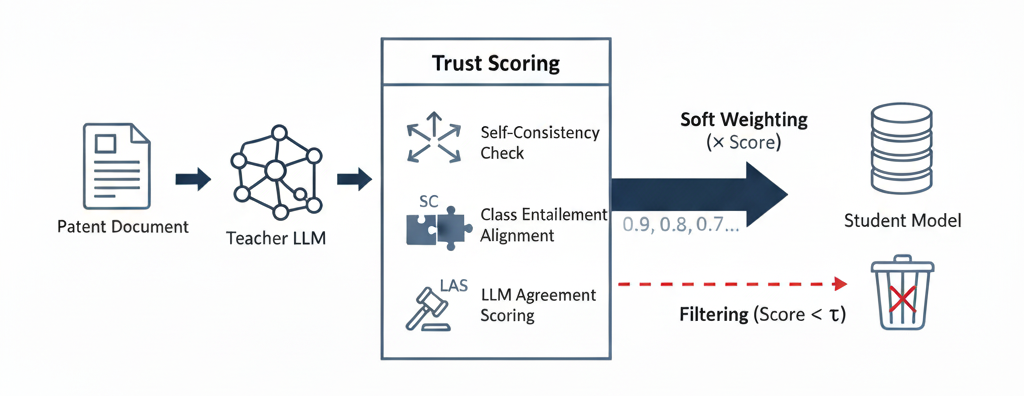}
  \caption{Overview of Self-Filtered Distillation (SFD). Trust indicators assess LLM-generated rationale quality, enabling selective distillation that ensures classification reliability for downstream patent information management.}
  \label{fig:intro_overview}
  \vspace{-4mm}
\end{figure}

Beyond predictive accuracy, research has increasingly emphasized explanation-driven learning as a means of enhancing interpretability and trustworthiness in information-intensive classification tasks. This line of work leverages human-annotated rationales as auxiliary supervision signals, demonstrating that explanations can improve both generalization and interpretability~\citep{camburu2018esnli,deyoung2020eraser}. While promising, such approaches incur high annotation costs and are often limited to short texts and simple classification tasks. Consequently, they fall short when applied to the classification of multifaceted and lengthy patents, where explanation-driven supervision faces domain-specific challenges.

With the advent of LLMs, research has shifted toward leveraging model-generated rationales to strengthen learning without requiring human annotation~\citep{wei2022chain,hsieh2023distilling}. While this paradigm offers scalability and interpretability, LLM-generated rationales often contain logical inconsistencies, label mismatches, and semantic misalignment with domain-specific classification schemes~\citep{turpin2023language} that are unsuitable for patents. For instance, an LLM may justify categorizing a digital communication patent (H04L) based on antenna structures or misclassify an artificial intelligence patent (G06N) as general data processing (G06F). Such errors extend beyond performance degradation: in the patent domain, they can lead to misguided technology mapping, flawed market analysis, or even intellectual property disputes. Moreover, misclassification directly compromises prior art retrieval, potentially causing examiners to overlook critical references and exposing granted patents to invalidity challenges. Treating all rationales as direct supervisory signals risks propagating these errors and reducing training stability. Therefore, high-stakes applications of patent classification demand a framework that can quantitatively assess rationale quality and selectively incorporate only trustworthy rationales~\citep{lampinen2022can}.

To address this challenge, \textbf{S}elf-\textbf{F}iltered \textbf{D}istillation (SFD) reinterprets LLM-generated rationales as trust indicators rather than ground-truth labels, dynamically weighting their contribution and filtering out extremely low-trust cases. We propose three unsupervised trust metrics to guide the distillation process: (1) Self-Consistency, which measures the stability of multiple rationales generated for the same input; (2) Class Entailment Alignment, which evaluates semantic coherence between rationales and class-specific definitions; and (3) LLM Agreement Scoring, which employs an external verifier LLM to assess the plausibility of rationale-label pairs. These metrics are aggregated into a unified trust score that dynamically modulates the contribution of each training instance during distillation, thereby ensuring that only reliable reasoning informs the classification and organization of patent knowledge.

Empirical evaluations on the USPTO-2M dataset demonstrate that SFD consistently surpasses traditional label-based learning and conventional distillation methods across diverse student architectures, achieving up to 38.7\% relative improvement in Macro-F1 with strong alignment between trust scores and expert judgments ($r=0.685$). These results confirm that unsupervised trust indicators can serve as effective quality filters, establishing a robust learning paradigm for leveraging LLM-generated rationales not only in patent classification but also in broader information-intensive domains where reliability and transparency are essential.

The primary contributions include:
\begin{itemize}
    \item \textbf{Trust-aware selective distillation}: We propose Self-Filtered Distillation, which reinterprets LLM-generated rationales as trust indicators and integrates three unsupervised metrics (Self-Consistency, Class Entailment Alignment, and LLM Agreement Scoring) to dynamically modulate training contributions, advancing knowledge distillation under noisy supervision.
    
    \item \textbf{Reliable information organization framework}: The trust score mechanism provides quantitative confidence measures for each classification decision, enabling auditability and transparent quality control over automated patent knowledge organization at scale.

    \item \textbf{High-stakes domain validation}: We validate SFD on patent classification, a domain where misclassification carries tangible consequences for prior art retrieval, technology valuation, and legal decision-making, demonstrating that trust-aware distillation is effective precisely where classification reliability is most critical.    
\end{itemize}

\section{Related Work}

\subsection{Patent Classification}

Patent classification supports core information management functions including prior art retrieval, technology mapping, and legal decision-making. The field has progressed from TF–IDF with SVMs~\citep{fall2003automated}, through CNN- and RNN-based models that capture semantic patterns in claims and abstracts~\citep{li2018deeppatent}, to transformer-based architectures that leverage contextual representations. Domain-adapted encoders such as PatentBERT~\citep{lee2020patent} and PatentSBERTa~\citep{bekamiri2021patentsberta} outperform general-purpose language models by pretraining on patent corpora, while generative approaches like PatentGPT~\citep{bai2024patentgpt} demonstrate that large-scale pretraining can further improve classification through richer language understanding. Exploiting the hierarchical structure of the Cooperative Patent Classification (CPC) system has also proven effective for managing the complexity of multi-label assignment~\citep{xu2021hierarchical}, and large-scale benchmarks such as USPTO-2M~\citep{risch2020patentmatch}, comprising millions of patents across hundreds of CPC subclasses, have become standard evaluation testbeds.

Despite these advances, patents remain fundamentally harder to classify than general-domain texts due to specialized terminology, lengthy claim structures, concept interdependencies, and severe label imbalance. More critically, existing approaches focus almost exclusively on predictive accuracy, leaving a gap in transparency and interpretability that is essential when classification outcomes directly inform downstream knowledge organization and legal decisions. This limitation motivates explanation-aware frameworks that can provide not only accurate labels but also auditable reasoning for each classification outcome.

\subsection{Explanation-driven Learning}
Chain-of-Thought (CoT) prompting enhances LLM performance and interpretability by generating intermediate rationales~\citep{wei2022chain, kojima2022zero, wang2023selfconsistency}, inspiring distillation methods where student models learn from teacher reasoning traces~\citep{hsieh2023distilling}. While approaches like KeyPoint-CoT improve efficiency~\citep{feng2024keypointcot}, most assume uniform rationale quality, risking error propagation when misaligned rationales distort label assignments. More recent work has begun to address this gap: QCRD~\citep{wang2025qcrd} employs self-consistency for denoising and trains an online discriminator to weight rationale quality during contrastive distillation, while FRODO~\citep{paul2024faithfulness} introduces causal objectives to improve faithfulness of generated reasoning steps. However, these methods either require additional learnable components or target general reasoning benchmarks, leaving the problem of rationale reliability in domain-specific classification largely unaddressed.

Despite these advances, LLM rationales remain fundamentally unreliable as supervision signals. Explanation-driven learning utilizes rationales as auxiliary supervision, with human-annotated datasets demonstrating improved generalization and robustness~\citep{camburu2018esnli, deyoung2020eraser, lampinen2022can, wiegreffe2021measuring}. Although metrics like ROSCOE exist to assess faithfulness~\citep{golovneva2023roscoe, hase2020evaluating}, LLM rationales often suffer from incompleteness and label inconsistencies~\citep{turpin2023language, ye2022unreliability, zhu2025rationales}. These flaws are amplified in patent classification due to its complex technical vocabulary and hierarchical categorization schemes, yet no existing method systematically quantifies rationale reliability to guard against such noise during training.


\subsection{Trustworthiness and Selective Learning}
Ensuring LLM trustworthiness is critical in decision-sensitive contexts. Existing work largely focuses on inference-time reliability through methods like Self-Consistency~\citep{wang2023selfconsistency} and semantic uncertainty quantification~\citep{kuhn2023semantic}. Other approaches employ selective prediction or self-evaluation for confidence scoring~\citep{yoshikawa2023selective, chen2023selfeval}. While uncertainty-aware fine-tuning helps mitigate hallucinations~\citep{krishnan2024enhancing, wang2024uncertainty, damani2025beyond}, these safeguards remain confined to inference or require specific tuning objectives. Research on self-improvement explores bootstrapping via synthetic data, often leveraging Reinforcement Learning (RL)~\citep{zelikman2022star, huang2023selfimprove}. However, these powerful methods typically rely on ground-truth verification or computationally expensive reward models to filter outputs. From a data-centric perspective, recent work emphasizes that the quality of training data is as important as model architecture~\citep{xu2024datacentric}, yet principled methods for assessing and leveraging the quality of LLM-generated rationales during training remain scarce.

Our work integrates trust indicators directly into the training process as a more efficient alternative. Instead of post-hoc filtering or RL loops, we dynamically reweight training instances based on rationale quality using Self-Consistency, Class Entailment Alignment, and LLM Agreement Scoring. This selective distillation strategy enables reliability-aware training without expensive annotations or additional learnable parameters. In high-stakes domains like patent classification, where misclassification propagates errors through downstream information management pipelines, this approach prevents error propagation from low-quality rationales, advancing beyond inference-level reliability toward responsible, stability-guided training.
\section{Methodology}

\begin{figure*}[t]
    \centering    
    \includegraphics[width=0.99\textwidth,keepaspectratio]{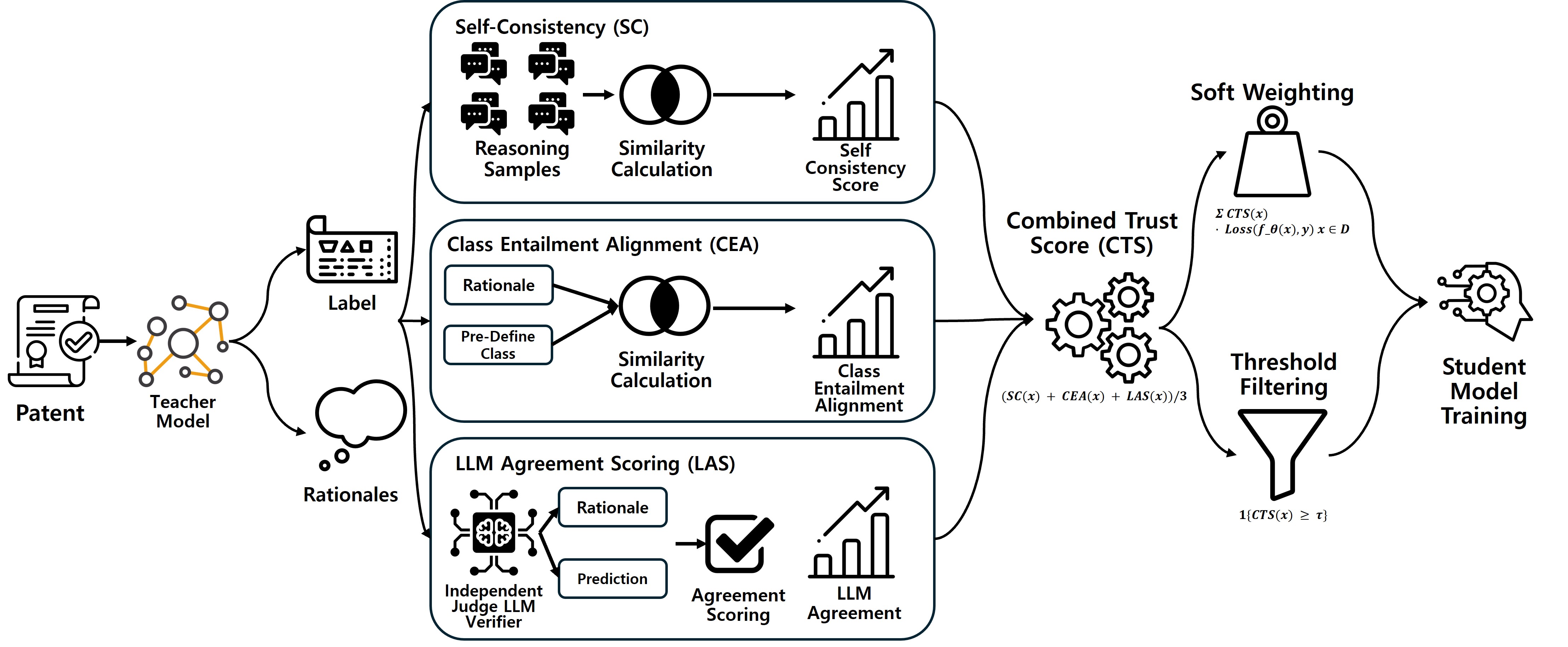}
    \caption{Overview of the proposed Self-Filtered Distillation (SFD) framework. It evaluates the quality of LLM-generated rationales using three unsupervised trust indicators (SC, CEA, and LAS) and applies trust-aware training control (soft weighting and threshold filtering) to optimize model training with high-quality samples.}
    \label{fig:SFD_workflow}
    \vspace{-4mm}
\end{figure*}

The Self-Filtered Distillation (SFD) framework, shown in Figure~\ref{fig:workflow}, quantitatively evaluates the quality of reasoning generated by LLMs and dynamically adjusts the learning contribution according to its estimated trustworthiness. The training process consists of three stages: Step 1 with LLM-based prediction and reasoning generation, Step 2 with reasoning quality evaluation, and Step 3 for trust-aware loss modulation.

\subsection{LLM-Based Prediction and Reasoning Generation}

In our setting, a teacher LLM receives an input patent text \( x \) and is prompted to generate two outputs: (1) a set of predicted labels \( \hat{Y} \subseteq \mathcal{Y} \), where \( \mathcal{Y} \) denotes the full label space, and (2) a natural language rationale \( r \) that explains the prediction. The predicted labels serve as soft supervision for the student model, while the generated rationale is not treated as a ground-truth signal but as auxiliary information for estimating sample reliability.

Formally, the student model \( f_\theta \) takes the same input \( x \) and produces a probability vector \( \hat{\mathbf{y}} = f_\theta(x) \in [0,1]^{|\mathcal{Y}|} \), where each component \( \hat{y}_c \) indicates the confidence for label \( c \in \mathcal{Y} \). The ground-truth label is denoted as \( y \). The student is trained to minimize a label-based loss using \( \hat{\mathbf{y}} \) against the teacher-provided labels. At inference time, thresholds or task-specific decision rules are applied to derive final predictions.



\subsection{Reasoning Quality Evaluation: Lightweight Trust Scoring}

Rather than serving as ground-truth supervision, LLM-generated reasoning is interpreted as a trust signal that reflects whether the reasoning is reliable enough to guide training. This enables a selective distillation mechanism: high-trust rationales contribute more to model updates, while noisy or inconsistent rationales are automatically downweighted.

To quantify the trustworthiness of rationales without relying on human annotations, we introduce three complementary unsupervised metrics, as shown in Figure~\ref{fig:SFD_workflow}. Each captures a distinct facet of rationale quality.

\paragraph{\textbf{(1) Self-Consistency Scoring (SC)}}  
To assess the logical stability of LLM-generated reasoning, we prompt the same model \( k \) times with identical input \( x \), obtaining a set of reasonings \( \{r_1, \dots, r_k\} \). The prompt used for this multi-sample generation is detailed in Appendix~\ref{app:sc_prompt}. We then compute the average pairwise cosine similarity between their sentence embeddings \( \phi(r_i) \):
\begin{equation}
\mathrm{SC}(x) = \frac{2}{k(k-1)} \sum_{i<j} \cos\left(\phi(r_i), \phi(r_j)\right).
\end{equation}

Higher SC values indicate that the model produces consistent rationales across runs, suggesting stable reasoning with reduced stochasticity.

For instance, when the same patent document describing a convolutional neural network for image classification produces divergent rationales across multiple generations, such as one emphasizing ``feature extraction through convolutional layers'' and another focusing on ``natural language tokenization for text processing,'' the resulting SC score is low. Such inconsistency indicates a lack of reasoning stability, which our framework interprets as reduced reliability during training.

\paragraph{\textbf{(2) Class Entailment Alignment (CEA)}}  
CEA measures how well the reasoning $r$ semantically aligns with each predicted label $\hat{y}$. For each predicted label, we query an LLM to obtain its class definition $d_{\hat{y}}$, and then compute the cosine similarity between the embeddings of $r$ and $d_{\hat{y}}$.  
\begin{equation}
\mathrm{CEA}(x) = \frac{1}{|\hat{Y}|} \sum_{\hat{y} \in \hat{Y}} \cos \bigl(\phi(r), \phi(d_{\hat{y}})\bigr).
\end{equation} 

By averaging over the predicted label set $\hat{Y}$, CEA evaluates the rationale's consistency with the label space. The Cooperative Patent Classification (CPC) system follows a hierarchical structure, where broader technical domains (e.g., G06F for computer technology) contain specific subclasses (e.g., G06N for artificial intelligence). CEA leverages this by extracting definitions of predicted CPC codes via an LLM and aligning them with the rationale. For multi-label cases, CEA averages alignment across all candidate definitions, ensuring coherence with the domain taxonomy. For instance, when a document is assigned to H04L (digital communication) but the rationale emphasizes ``antenna structure'' rather than ``signal transmission protocols,'' the semantic mismatch results in a low CEA score, indicating reduced trustworthiness.

\paragraph{\textbf{(3) LLM Agreement Scoring (LAS)}}  
LAS assesses the external plausibility of reasoning using an independent LLM-based verifier $\mathrm{LLM}_{\mathrm{judge}}$, separate from the model that generates the reasoning. Generative models often exhibit overconfidence in their own outputs, which can lead to unreliable rationales if left unchecked. To mitigate this issue, the verifier receives a reasoning–label pair $(r, \hat{y})$ and outputs a scalar confidence score. The detailed judging prompt is provided in Appendix~\ref{app:las_prompt}.
\begin{equation}
\mathrm{LAS}(x) = \sigma\left(\mathrm{LLM}_{\mathrm{judge}}(r, \hat{y})\right).
\end{equation}

Here, $\sigma$ is the sigmoid function mapping the raw score into $[0, 1]$. By introducing an external verifier, LAS provides an additional validation layer that counteracts model overconfidence and complements the other trust metrics by evaluating plausibility from an independent perspective.  

For instance, in cases where the original text concerns ``data encryption methods'' but the generated rationale incorrectly frames the invention as addressing ``data compression,'' the external verifier assigns a low LAS score. This penalization captures the lack of plausibility in the rationale–label alignment and highlights the necessity of incorporating independent validation to mitigate overconfidence in model-generated explanations.

SC, CEA, and LAS capture complementary facets of rationale quality: internal stability, taxonomic alignment, and external plausibility, respectively. By combining these perspectives into a single score without additional learnable parameters, our framework avoids the computational overhead of training a separate discriminator while providing richer quality signals than any single-indicator approach. This multi-faceted design enables the detection of failure modes that would escape methods relying on consistency alone or external verification alone.

\subsection{Trust-Aware Training Control}

The Combined Trust Score (CTS) derived from our multi-aspect evaluation is used to modulate the learning contribution of each training sample. Our framework adopts a soft weighting strategy where all samples participate in training, but their loss contributions are scaled proportionally to their estimated trustworthiness. For instance, a sample with a low trust score, $\mathrm{CTS}(x)\approx 0.5$, has its influence significantly reduced. This approach mitigates the negative impact of low-quality reasoning without entirely discarding potentially informative signals.

Formally, let the model $f_\theta$ predict a probability vector $\hat{\mathbf{y}} = f_\theta(x) \in [0,1]^{|\mathcal{Y}|}$ for an input $x \in \mathcal{X}$, with the ground-truth label denoted as $y \in \mathcal{Y}$. The task-specific loss function, $\mathrm{Loss}(\cdot)$ (e.g., cross-entropy for classification), is scaled by the trust score $\mathrm{CTS}(x)\in[0,1]$.

The final trust score, referred to as CTS, is computed as the direct average of the three previously defined metrics (SC, CEA, and LAS). This weighting scheme treats each dimension of trustworthiness as equally important, providing a balanced and robust measure of rationale quality without introducing additional learnable parameters. The CTS is thus defined as:
\begin{equation}
    \mathrm{CTS}(x)= \frac{1}{3}\big(\mathrm{SC}(x) + \mathrm{CEA}(x) + \mathrm{LAS}(x)\big).
\end{equation}

The overall training objective is to minimize the trust-weighted loss across all samples in data $\mathcal{D}$:
\begin{equation}
    \mathcal{L}= \sum_{x\in \mathcal{D}} \mathrm{CTS}(x)\cdot \mathrm{Loss}\big(f_{\theta}(x),\, y\big).
\end{equation}

In addition to this soft weighting, we introduce a thresholding mechanism to completely exclude samples with extremely low trust scores. An indicator function, $\mathbf{1}\{\cdot\}$, is applied to filter out any sample whose trust score falls below a predefined threshold $\tau$:
\begin{equation}
    \mathcal{L}= \sum_{x\in \mathcal{D}} \mathbf{1}\{\mathrm{CTS}(x)\ge \tau\}\cdot \mathrm{CTS}(x)\cdot \mathrm{Loss}\big(f_{\theta}(x), y\big).
\end{equation}

Candidate thresholds, such as $\tau\in\{0.1, 0.2, 0.3, \ldots\}$, are evaluated on a validation set, and the optimal value $\tau^\ast$ is selected for the final training. This strategy of combining soft weighting and hard thresholding ensures that the model remains robust against noisy rationales while effectively leveraging high-quality reasoning signals.

\section{Experiments and Results}

We evaluate the proposed framework by addressing five research questions that collectively assess its effectiveness, robustness, and interpretability in patent classification. 
\textbf{RQ1}: How does SFD compare with baseline methods? \textbf{RQ2}: Does threshold-based filtering improve robustness compared to soft weighting? \textbf{RQ3}: What is the effect of removing each trust metric? \textbf{RQ4}: Is the proposed framework model-agnostic across different student architectures? \textbf{RQ5}: Are trust scores correlated with human judgments?

\subsection{Dataset}
\label{sec:dataset}
We employ the USPTO-2M dataset~\citep{li2018deeppatent}, a predominant benchmark in multi-label patent classification comprising 2,000,147 US utility patents filed between 2006 and 2015. The dataset is split into 1,950,247 training documents and 49,900 test documents. To balance coverage with computational feasibility, we restrict the label space to the CPC subclass level (4-character codes, e.g., G06F, H04L), resulting in 637 candidate labels in the training set, a standard practice in prior studies~\citep{lee2020patent,patentnet}. The dataset exhibits severe label imbalance across subclasses, making Macro-F1 a particularly informative evaluation metric alongside Micro-F1. USPTO-2M serves as a rigorous foundation for evaluating classification performance given the inherent challenges of patent texts, including specialized terminology, complex claim structures, and concept interdependencies that distinguish them from general-domain documents. Subset Accuracy measures the fraction of instances for which the predicted label set exactly matches the ground truth.

\subsection{Experimental Settings}

We employ Qwen3-30B as the teacher and validate our framework across four diverse student models: Qwen3-0.6B, Phi-4-Mini, Llama-3.2-3B, and Falcon3-3B. Rationales are generated via Nucleus sampling ($T=0.6, p=0.9$), with results averaged over five independent runs to ensure reliability. For trust evaluation, we sample three rationales per instance and set equal coefficients ($\lambda=1/3$). We utilize Qwen3-32B as the judge for its strong legal reasoning capabilities and all-MiniLM-L6-v2 for embeddings. Training uses the Adam optimizer with a learning rate of $1\times10^{-5}$. All experiments are conducted on NVIDIA H100 GPUs. All code and checkpoints will be released upon acceptance.

\subsection{RQ1. How does SFD compare with baseline methods?}

We compare SFD against five distillation baselines that span a spectrum of rationale utilization strategies. To ensure a fair comparison, all methods use the identical student architecture (Qwen3-0.6B) and training protocol (same optimizer, learning rate, and number of training steps), isolating the effect of each distillation strategy from model capacity.

\paragraph{Baseline descriptions.}
\textbf{Label-only Distillation} trains the student using only the teacher's predicted labels without any rationale information, serving as the lower-bound reference.
\textbf{Naive CoT Distillation} treats all teacher-generated rationales as ground-truth supervision and appends them to the input during training, following the standard distilling step-by-step paradigm~\citep{hsieh2023distilling}.
\textbf{Self-Consistency Distillation} generates five rationales per instance via independent sampling, retains only those whose predicted labels agree with the majority vote, and discards the rest. This mirrors the self-consistency selection mechanism of~\citet{wang2023selfconsistency}.
\textbf{LLM-as-a-Judge Distillation} employs an external verifier (Qwen3-32B) to score each rationale on a 1--10 scale; rationales scoring below 5 are discarded. This follows the single-judge filtering paradigm.
\textbf{Rationale-Augmented Distillation} concatenates all generated rationales as additional input features to the student encoder without any quality filtering, testing whether raw rationale information alone provides benefit.

\begin{table}[ht]
\centering
\footnotesize
\resizebox{0.95\linewidth}{!}{%
\begin{tabular}{lcc}
\toprule
\textbf{Method} & \textbf{F1-Micro} & \textbf{F1-Macro} \\
\midrule
\multicolumn{3}{l}{\textbf{\textit{Baseline Methods (Qwen3-0.6B)}}} \\
Label-only Distillation          & 0.821 & 0.401 \\
Naive CoT Distillation           & 0.942 & 0.401 \\
Self-Consistency Distillation    & 0.932 & 0.354 \\
LLM-as-a-Judge Distillation      & 0.893 & 0.355 \\
Rationale-Augmented Distillation & 0.944 & 0.369 \\
\midrule
\multicolumn{3}{l}{\textbf{\textit{Our Framework (SFD, Qwen3-0.6B)}}} \\
\textbf{SFD}       & \textbf{0.960} & \textbf{0.404} \\
\bottomrule
\end{tabular}
}
\caption{Performance comparison on USPTO-2M. All methods use Qwen3-0.6B as the student model for fair comparison. SFD achieves the highest Micro-F1 while maintaining competitive Macro-F1. Results with larger student architectures are reported in Table~\ref{tab:rq4-model-agnostic}.}
\label{tab:main_results}
\end{table}

\paragraph{\textbf{Results.}}
As shown in Table~\ref{tab:main_results}, SFD achieves the highest Micro-F1 (0.960), surpassing the strongest baseline (Rationale-Augmented Distillation, 0.944) by 1.6 percentage points. An important observation emerges from the single-indicator baselines: both Self-Consistency Distillation (0.932) and LLM-as-a-Judge Distillation (0.893) underperform Naive CoT Distillation (0.942), which applies no filtering at all. This counterintuitive result suggests that relying on a single quality signal can introduce harmful filtering bias, discarding useful rationales while retaining harmful ones that happen to pass the narrow criterion. In contrast, SFD's multi-faceted trust score integrates complementary perspectives (internal stability, taxonomic alignment, and external plausibility), enabling more accurate discrimination between reliable and unreliable rationales.

\paragraph{\textbf{Robustness and statistical significance.}}
We conducted five independent runs for all methods. SFD achieved $0.960 \pm 0.006$ Micro-F1, confirming stable performance with minimal variance across random seeds. All pairwise differences between SFD and each baseline in Micro-F1 are statistically significant (paired $t$-test, $p < 0.05$).

We note that the Macro-F1 gap in Table~\ref{tab:main_results} is narrow (0.404 vs.\ 0.401). This is a direct consequence of the aggressive filtering threshold ($\tau > 0.9$), which prioritizes precision by retaining only the highest-quality rationales. Because minority classes have fewer training instances to begin with, strict filtering disproportionately reduces their effective training set, limiting Macro-F1 gains at the 0.6B scale. Crucially, this is not a limitation of the trust mechanism itself but of the student model's capacity to generalize from a reduced sample. When paired with larger student architectures that possess sufficient representational power to learn from fewer but cleaner examples, SFD yields Macro-F1 of 0.556 (Falcon3-3B), a 38.7\% relative improvement over the same baseline (Table~\ref{tab:rq4-model-agnostic}). We interpret Macro-F1 improvement as a capacity-dependent benefit that the trust mechanism enables but does not solely determine.

\subsection{RQ2. Does threshold-based filtering improve robustness compared to soft weighting?}

\begin{table}[ht]
\centering
\footnotesize
\resizebox{0.9\linewidth}{!}{%
\begin{tabular}{lccc}
\toprule
\textbf{Threshold ($\tau$)} & {\textbf{F1-Micro}} & {\textbf{F1-Macro}} & \textbf{Subset Acc.} \\
\midrule
No Trust Weighting & 0.821 & 0.401 & 0.648 \\
$\tau > 0.5$ & 0.820 & 0.400 & 0.646 \\
$\tau > 0.6$ & 0.828 & 0.421 & 0.660 \\
$\tau > 0.7$ & 0.837 & 0.439 & 0.677 \\
$\tau > 0.8$ & 0.890 & \textbf{0.462} & 0.793 \\
$\tau > 0.9$ & \textbf{0.960} & 0.404 & \textbf{0.930} \\
\bottomrule
\end{tabular}
}
\caption{Effect of threshold-based filtering using Qwen3-0.6B on USPTO-2M. ``No Trust Weighting'' denotes uniform training without trust scores. Progressively stricter filtering improves robustness, with $\tau > 0.9$ achieving peak Subset Accuracy of 0.930.}
\label{threshold_analysis}
\end{table}

We investigate how aggressively low-trust samples should be excluded during training. Using Qwen3-0.6B as the student model, we compare uniform training without trust scores against threshold-based filtering at five granularity levels ($\tau \in \{0.5, 0.6, 0.7, 0.8, 0.9\}$).

Table~\ref{threshold_analysis} reveals a clear trend: as the threshold increases, both Micro-F1 and Subset Accuracy improve monotonically. At the lenient end ($\tau > 0.5$), performance is virtually identical to uniform training, indicating that low-confidence samples below 0.5 are rare and their removal has negligible effect. In contrast, raising the threshold to $\tau > 0.9$ produces a sharp jump, achieving Micro-F1 of 0.960 and Subset Accuracy of 0.930. This demonstrates that a substantial portion of rationales scored between 0.5 and 0.9 still carry noise that degrades training when included indiscriminately.

However, a trade-off emerges in Macro-F1. The best Macro-F1 (0.462) occurs at $\tau > 0.8$, whereas $\tau > 0.9$ reduces it to 0.404. This suggests that the strictest threshold discards some genuinely informative samples from minority classes, where fewer training instances are available and even moderately scored rationales contribute useful signal.

We adopt $\tau > 0.9$ as the default configuration for subsequent experiments, prioritizing overall classification reliability (Micro-F1 and Subset Accuracy). The Macro-F1 reduction is recoverable: as shown in Table~\ref{tab:rq4-model-agnostic}, larger student models with greater representational capacity restore strong Macro-F1 performance (up to 0.556) by better leveraging the high-quality rationales that pass the strict filter.
\subsection{RQ3. What is the effect of removing each trust metric?}



\begin{table}[ht]
\centering
\footnotesize  
\resizebox{0.9\columnwidth}{!}{%
\begin{tabular}{lccc}
\toprule
Method & F1-Micro & F1-Macro & Subset Acc. \\
\midrule
SC only  & 0.932 & 0.354 & 0.898 \\
CEA only & 0.952 & 0.367 & 0.922 \\
LAS only & 0.893 & 0.355 & 0.844 \\
\midrule
w/o LAS  & 0.922 & 0.346 & 0.859 \\
w/o CEA  & 0.797 & 0.262 & 0.695 \\
w/o SC   & 0.957 & 0.359 & 0.930 \\
\midrule
\textbf{$\mathrm{CTS}(x)$} & \textbf{0.960} & \textbf{0.404} & \textbf{0.930} \\
\bottomrule
\end{tabular}
}
\caption{Ablation study of trust metrics on USPTO-2M ($\tau > 0.9$), reporting Micro-F1, Macro-F1, and Subset Accuracy. SC = Self-Consistency, CEA = Class Entailment Alignment, LAS = LLM Agreement Scoring.}
\label{tab:rq3-ablation}
\end{table}

To understand the individual contribution of each trust metric, we conduct an ablation study using Qwen3-0.6B with the threshold fixed at $\tau > 0.9$. Table~\ref{tab:rq3-ablation} reports results under three conditions: using each metric in isolation (upper block), removing each metric from the full score (middle block), and the complete integrated score CTS (bottom row).

The most striking finding is the critical role of CEA. When CEA is removed, Subset Accuracy collapses from 0.930 to 0.695, and Macro-F1 drops to 0.262. This severe degradation indicates that without explicit alignment to CPC class definitions, the trust score loses its ability to distinguish domain-relevant reasoning from superficially plausible but taxonomically misaligned rationales. Conversely, CEA alone achieves the strongest single-metric performance (Micro-F1 0.952, Subset Accuracy 0.922), confirming that taxonomic grounding is the most informative individual signal for patent classification.

LAS contributes a complementary filtering function. Removing LAS causes Micro-F1 to fall from 0.960 to 0.922 and Subset Accuracy from 0.930 to 0.859. While CEA ensures taxonomic relevance, LAS identifies rationales that appear semantically coherent yet contain factual hallucinations detectable only through external verification. The combination of CEA and LAS thus provides both domain grounding and factual validation.

The effect of SC is more subtle. Removing SC preserves Subset Accuracy at 0.930 and only marginally reduces Micro-F1 (0.957), but lowers Macro-F1 from 0.404 to 0.359, an 11.1\% relative drop. This pattern indicates that self-consistency primarily benefits minority classes, where the LLM tends to produce more variable rationales due to limited exposure during pre-training. For high-frequency classes with abundant training signal, the other two metrics suffice; for rare classes, SC provides an essential stability check.

In summary, the three metrics address distinct failure modes: CEA guards against taxonomic drift, LAS filters hallucinated reasoning, and SC stabilizes predictions for underrepresented categories. Their integration into a unified CTS yields the strongest overall performance, confirming that no single perspective is sufficient for reliable trust estimation in patent classification.

\subsection{RQ4. Is the proposed framework model-agnostic across different student architectures?}

\begin{table}[ht]
\centering
\footnotesize
\resizebox{0.95\linewidth}{!}{%
\begin{tabular}{lrccc}
\toprule
\textbf{Student Model} & \textbf{Params} & \textbf{F1-Micro} & \textbf{F1-Macro} & \textbf{Subset Acc.} \\
\midrule
Qwen3-0.6B   & 0.6B & 0.9600 & 0.4040 & 0.9330 \\
Phi-4-Mini   & 3.8B & 0.9685 & 0.5025 & 0.9453 \\
Llama-3.2-3B & 3.2B & 0.9880 & 0.5153 & 0.9765 \\
\textbf{Falcon3-3B}   & 3.0B & \textbf{0.9881} & \textbf{0.5556} & \textbf{0.9766} \\
\bottomrule
\end{tabular}
}
\caption{SFD performance across diverse student architectures on USPTO-2M. All configurations use the same trust-aware distillation pipeline.}
\label{tab:rq4-model-agnostic}
\end{table}

A practical distillation framework must generalize beyond a single student architecture. To test this, we apply the identical SFD pipeline (same teacher, same trust threshold $\tau > 0.9$, same training protocol) to four student models drawn from distinct model families and parameter scales: Qwen3-0.6B (0.6B), Phi-4-Mini (3.8B), Llama-3.2-3B (3.2B), and Falcon3-3B (3.0B).

Table~\ref{tab:rq4-model-agnostic} shows that SFD delivers consistently strong results regardless of the underlying architecture. All four models exceed 0.960 Micro-F1, demonstrating that trust-aware filtering reliably removes harmful rationales irrespective of how the student encodes and processes them. This consistency across architectures with fundamentally different tokenizers, pre-training corpora, and attention mechanisms confirms that the trust score captures rationale quality at a level independent of student-side inductive biases.

Beyond this baseline consistency, a clear capacity effect emerges in Macro-F1. Qwen3-0.6B achieves 0.404, while the three 3B-scale models reach 0.503 to 0.556. The best-performing model, Falcon3-3B, attains a Macro-F1 of 0.556, representing a 38.7\% relative improvement over the strongest baseline in Table~\ref{tab:main_results} (0.401). This gap is primarily attributable to minority-class performance: larger models have sufficient representational capacity to learn meaningful decision boundaries even from the reduced training set that survives strict filtering, whereas the 0.6B model lacks the parameters to fully exploit these high-quality samples for rare categories.

Importantly, this does not diminish the contribution of the trust mechanism itself. Even at the 0.6B scale, SFD already outperforms all baselines in Micro-F1 (0.960 vs. 0.944), confirming that the trust score is the primary driver of improvement. Increased model capacity amplifies this benefit but does not create it. In practice, this means that practitioners can select their student model based on deployment constraints (latency, memory, cost) while retaining the full advantages of trust-aware distillation.

\subsection{RQ5. Are trust scores correlated with human judgments of reasoning quality?}

\begin{table}[ht]
\centering
\footnotesize
\resizebox{0.85\linewidth}{!}{%
\begin{tabular}{lccc}
\toprule
\textbf{Metric} & \textbf{Pearson ($r$)} & \textbf{$\Delta$} & \textbf{Rel. $\Delta$} \\
\midrule
SC only & 0.623 & $-$0.062 & $-$9.05\% \\
CEA only & 0.612 & $-$0.073 & $-$10.66\% \\
LAS only & 0.528 & $-$0.157 & $-$22.92\% \\
w/o LAS & 0.584 & $-$0.101 & $-$14.75\% \\
w/o CEA & 0.475 & $-$0.210 & $-$30.66\% \\
w/o SC & 0.631 & $-$0.054 & $-$7.88\% \\
\textbf{$\mathrm{CTS}(x)$} & \textbf{0.685} & \textbf{--} & \textbf{--} \\
\bottomrule
\end{tabular}
}
\caption{Pearson correlation ($r$) between trust scores and human judgments on 1,000 samples of USPTO-2M. $\Delta$ and Rel.~$\Delta$ denote absolute and relative differences from the full CTS.}
\label{tab:rq5-human-correlation}
\end{table}

Automated trust scores are only meaningful if they reflect genuine reasoning quality as perceived by domain experts. To validate this, we sampled 1,000 instances from USPTO-2M stratified across the full range of trust scores (low, medium, high) to ensure balanced coverage. Three annotators with patent examination experience independently rated each rationale on a 1--5 Likert scale along three dimensions: logical consistency, task alignment, and plausibility. Inter-annotator agreement reached Krippendorff's $\alpha = 0.641$, indicating moderate agreement that is consistent with the inherent subjectivity of rationale quality assessment~\citep{krippendorff2011computing}. Detailed annotation guidelines are provided in Appendix~\ref{app:annotation}.

Table~\ref{tab:rq5-human-correlation} reports the Pearson correlation between each trust metric configuration and the averaged human scores. The integrated $\mathrm{CTS}(x)$ achieves the highest correlation ($r = 0.685$), substantially outperforming any individual metric used in isolation. This confirms that the multi-faceted composition captures aspects of reasoning quality that no single signal can represent alone.

The ablation pattern mirrors the findings of RQ3. Removing CEA produces the largest degradation ($-30.66\%$), reinforcing its role as the dominant quality signal: because CEA anchors evaluation to objective CPC ontology definitions rather than model-internal heuristics, it aligns naturally with how human experts assess whether a rationale correctly addresses the assigned technical category. LAS removal causes the second-largest drop ($-22.92\%$), indicating that external plausibility verification captures a distinct quality dimension that humans also penalize when absent. SC removal has the smallest impact ($-7.88\%$), consistent with its role as a stability regularizer rather than a primary quality indicator.

An important implication of these results is that the trust score is not merely an internal training signal but also an interpretable and auditable output. In practical patent information management workflows, the trust score can serve as a confidence indicator attached to each classification decision, enabling downstream systems or human reviewers to prioritize cases where automated reasoning may be unreliable. The strong correlation with expert judgments ($r = 0.685$) provides empirical justification for this use case, bridging the gap between automated scalability and the transparency requirements of high-stakes intellectual property decisions.

\section{Discussion}

\subsection{Why Does Multi-Faceted Trust Scoring Work?}

The counterintuitive failure of single-indicator baselines (RQ1) reveals a fundamental insight: rationale quality is not a unidimensional construct. A rationale can be internally consistent yet taxonomically misaligned (escaping SC but caught by CEA), or taxonomically plausible yet factually hallucinated (escaping CEA but caught by LAS). This observation challenges the implicit assumption in prior distillation work that a single quality signal can serve as a reliable gatekeeper for training data.

We argue that the effectiveness of CTS stems from its coverage of orthogonal failure modes rather than from any individual metric's superiority. The ablation results (RQ3) provide empirical support: CEA alone achieves 0.952 Micro-F1, yet integrating SC and LAS still yields a meaningful gain to 0.960. More critically, the human correlation study (RQ5) shows that removing any single component degrades alignment with expert judgments, with the degradation patterns mirroring the performance ablation. This convergence between task performance and human perception suggests that our three metrics approximate genuinely distinct dimensions of reasoning quality rather than redundantly measuring the same underlying signal.

This finding carries implications beyond patent classification. In any domain where LLM-generated explanations guide downstream decisions, the reliability of those explanations is unlikely to be captured by a single heuristic. The principle of triangulating quality through complementary, independently computed signals suggests a transferable design principle for trustworthy knowledge distillation. Rather than asking ``is this rationale good?'' through one lens, the multi-faceted approach asks ``does this rationale survive scrutiny from multiple independent perspectives?'', a standard that mirrors how human experts themselves evaluate reasoning quality by cross-referencing internal logic, domain knowledge, and external evidence.

\subsection{Rethinking Quality Assurance in Automated Knowledge Organization}

Large-scale knowledge organization systems have long relied on a sequential pipeline: first automate classification, then verify outputs through post-hoc quality control. This paradigm assumes that errors can be detected and corrected after they enter the system. However, as the volume of digital knowledge artifacts grows (4.5 million patent applications annually, millions of scientific publications, legal documents, and clinical records), post-hoc verification becomes a bottleneck that fundamentally limits the scalability of knowledge organization infrastructure.

Our work proposes a paradigm shift: embedding quality assurance directly into the learning process itself, so that the resulting model is trained to be reliable rather than corrected after the fact. The trust score does not merely filter training data; it encodes a quantitative theory of what constitutes trustworthy automated reasoning within a classification taxonomy. This reframing transforms quality control from an external auditing function into an intrinsic property of the learned model, a transition from ``classify first, verify later'' to ``learn only from what is verifiable.''

This shift addresses a core tension in information management research: the trade-off between automation scale and organizational reliability. Traditional approaches resolve this tension by accepting degraded quality at scale or limiting automation to high-confidence cases. SFD offers a third path, achieving both scale and reliability by ensuring that the training signal itself is curated for trustworthiness. The result is not merely a better classifier but a classification system whose outputs carry built-in confidence semantics, enabling downstream knowledge organization processes (prior-art retrieval, technology mapping, portfolio analysis) to propagate reliability guarantees rather than blindly inheriting upstream uncertainty.

The decomposability of CTS into three interpretable signals further advances this vision. When a classification decision receives high SC and LAS but low CEA, the system communicates that reasoning is internally stable and externally validated but potentially misaligned with the classification taxonomy, directing reviewer attention precisely where it is needed. This diagnostic granularity transforms automated classification from a take-it-or-leave-it prediction into a structured knowledge artifact that documents its own evidential basis. For information management systems that must maintain long-term consistency across millions of documents and evolving taxonomies, this self-documenting property represents a qualitative advance over opaque confidence scores.

\subsection{Scalability and Deployment Considerations}

The model-agnostic nature of SFD (RQ4) establishes a practical deployment principle: trust-aware distillation decouples quality control from model selection. Organizations can independently choose their student model based on infrastructure constraints (latency, memory, cost) and their trust threshold based on risk tolerance, without redesigning the distillation pipeline.

The computational overhead of trust scoring merits explicit discussion. Computing CTS requires three additional operations per training instance: generating three rationales (for SC), computing one embedding similarity (for CEA), and one external verifier call (for LAS). These costs are incurred once during data preparation and amortized across all subsequent training runs, regardless of how many student models are trained. In a production setting where the same patent corpus is distilled into multiple specialized models (e.g., by technology domain or deployment region), the trust scores become a reusable asset whose per-model cost decreases with scale.

The threshold $\tau$ offers an intuitive control surface for practitioners. Our results demonstrate a clear monotonic relationship between $\tau$ and classification reliability (RQ2), enabling organizations to calibrate the precision-coverage trade-off based on their specific operational requirements. Patent offices prioritizing examination accuracy may adopt $\tau > 0.9$, while technology scouting applications tolerating moderate noise may prefer $\tau > 0.7$ to retain broader coverage. This configurability, combined with the model-agnostic architecture, positions SFD as an adaptable infrastructure component for diverse patent information management scenarios rather than a fixed-configuration research prototype.

\section{Conclusion}

We presented Self-Filtered Distillation (SFD), a framework that reinterprets LLM-generated rationales as trust indicators rather than ground-truth supervision for large-scale patent knowledge organization. By integrating three unsupervised signals, Self-Consistency, Class Entailment Alignment, and LLM Agreement Scoring, into a unified trust score, SFD dynamically modulates training contributions and filters unreliable reasoning before it can propagate into the student model. Evaluated on the USPTO-2M benchmark comprising over two million patents across 637 CPC subclasses, SFD consistently outperforms existing distillation methods across four student architectures, achieving up to 38.7\% relative improvement in Macro-F1. The strong correlation between trust scores and expert judgments ($r = 0.685$) further demonstrates that the framework provides not only accurate but also interpretable and auditable classification outcomes.

Beyond predictive gains, this work contributes a broader methodological principle to information management research: shifting quality assurance from post-hoc verification of system outputs to intrinsic reliability embedded within the learning process itself. By attaching decomposable confidence semantics to each classification decision, SFD transforms automated patent classification from an opaque prediction task into a self-documenting knowledge organization process, reconciling the scalability demands of modern patent corpora with the transparency requirements of high-stakes intellectual property decisions.

Future directions include adaptive threshold learning to optimize minority-class performance, dynamic trust score updating as classification taxonomies evolve, and extension to other information-intensive domains such as legal document categorization, clinical coding, and academic literature organization, where reliable and auditable automated reasoning is equally essential.


\bibliographystyle{ACM-Reference-Format}
\bibliography{sample-base}


\appendix

\lstdefinestyle{promptcompact}{
  backgroundcolor=\color{promptbg},
  basicstyle=\ttfamily\scriptsize,
  breaklines=true,
  frame=single,
  numbers=none,
  xleftmargin=0.5em,
  xrightmargin=0.5em,
  aboveskip=4pt,
  belowskip=4pt,
}

\section{Annotation Guidelines}
\label{app:annotation}

To systematically evaluate the quality of LLM-generated rationales, we recruited three annotators and provided them with detailed guidelines to ensure consistency. Each rationale was evaluated on a 1--5 Likert scale according to three criteria: Logical Consistency, Task Alignment, and Plausibility. Below we describe the definitions, scoring criteria, and illustrative examples for each criterion.

\subsection{Logical Consistency}

\textbf{Definition:} Whether the rationale maintains internal consistency without contradictions.

\begin{table}[ht]
\centering
\scriptsize
\setlength{\tabcolsep}{2pt}
\begin{tabular}{cp{2.8cm}p{3.6cm}}
\toprule
\textbf{Score} & \textbf{Description} & \textbf{Example} \\
\midrule
1 & Severe contradictions undermining the explanation. & States ``implemented in hardware circuits'' but later claims ``operates purely as software.'' \\
2 & Multiple inconsistencies; illogical flow. & Key concepts appear disconnected and the reasoning is incoherent. \\
3 & Minor inconsistencies; overall acceptable. & Early part emphasizes classification algorithms, later shifts to preprocessing. \\
4 & Mostly consistent; small logical leaps. & Coherent around CNN-based classification but ends with a tangential remark. \\
5 & Completely consistent; clear logical flow. & Clearly explains CNN feature extraction and softmax classification without contradictions. \\
\bottomrule
\end{tabular}
\caption{Scoring rubric for Logical Consistency.}
\label{tab:logical_consistency}
\end{table}

\subsection{Task Alignment}

\textbf{Definition:} The extent to which the rationale aligns with the predicted CPC code definitions.

\begin{table}[ht]
\centering
\scriptsize
\setlength{\tabcolsep}{2pt}
\begin{tabular}{cp{2.8cm}p{3.6cm}}
\toprule
\textbf{Score} & \textbf{Description} & \textbf{Example} \\
\midrule
1 & Almost entirely unrelated to the CPC code. & For H04L (digital communication), discusses ``pharmaceutical compositions.'' \\
2 & Largely misaligned; irrelevant aspects. & For H04L, only describes ``antenna structure'' without addressing communication protocols. \\
3 & Generally aligned but missing key elements. & For H04L, discusses signal processing but omits transmission protocol. \\
4 & Well aligned; minor irrelevant details. & For H04L, primarily covers protocols and error correction, with slight mention of antenna design. \\
5 & Perfectly aligned with CPC definitions. & For H04L, explains ``data transmission protocols and error correction methods.'' \\
\bottomrule
\end{tabular}
\caption{Scoring rubric for Task Alignment.}
\label{tab:task_alignment}
\end{table}

\subsection{Plausibility}

\textbf{Definition:} Whether the rationale is technically factual and reasonable within the patent domain.

\begin{table}[ht]
\centering
\scriptsize
\setlength{\tabcolsep}{2pt}
\begin{tabular}{cp{2.8cm}p{3.6cm}}
\toprule
\textbf{Score} & \textbf{Description} & \textbf{Example} \\
\midrule
1 & Clearly unrealistic or factually incorrect. & Claims ``images are classified using tokenization techniques.'' \\
2 & Implausible or largely inaccurate. & ``An encryption patent drastically improves signal transmission speed.'' \\
3 & Exaggerated or incomplete but partially plausible. & ``The neural network achieves nearly perfect accuracy in classification.'' \\
4 & Mostly factual with minor inaccuracies. & ``CNN extracts image features and applies additional rule-based filtering.'' \\
5 & Fully factual, technically sound, and domain-appropriate. & ``CNN extracts features and softmax computes class probabilities.'' \\
\bottomrule
\end{tabular}
\caption{Scoring rubric for Plausibility.}
\label{tab:plausibility}
\end{table}

\subsection{Annotation Procedure}

\begin{itemize}
  \item Each annotator independently evaluated all 1{,}000 samples.
  \item For each rationale, scores were assigned for Logical Consistency, Task Alignment, and Plausibility, each on a 1--5 scale.
  \item The final human score for each sample was computed as the average across annotators.
  \item Inter-annotator reliability was measured with Krippendorff's alpha, yielding $\alpha = 0.641$, indicating acceptable agreement.
\end{itemize}

\section{Prompt for Class Entailment Alignment (CEA)}
\label{app:cea_prompt}

\begin{lstlisting}[style=promptcompact, caption={Prompt for Class Entailment Alignment (CEA).}, label={lst:cea-prompt}]
prompt_template = """
You are an expert on the Cooperative Patent
Classification (CPC) system.

Your task is to provide a concise, official
definition for the given CPC subclass code.
The definition should be a single, clear
sentence. Do not add any extra explanation
or introductory text.

CPC Code: {cpc_code}
Definition: """
\end{lstlisting}

\section{Prompt for LLM Agreement Scoring (LAS)}
\label{app:las_prompt}

\begin{lstlisting}[style=promptcompact, caption={Prompt for LLM Agreement Scoring (LAS).}, label={lst:las-prompt}]
LAS_PROMPT_TEMPLATE_CONTINUOUS = """
You are an impartial and strict judge. Your
task is to evaluate if the provided 'Reasoning'
logically and accurately supports the assignment
of the given 'Predicted Labels' based on the
'Original Text'.

Rate the quality of the reasoning on a scale
of 1 to 5.
- 1: Completely illogical, irrelevant, or
     hallucinatory.
- 2: Poorly reasoned, contains significant
     flaws.
- 3: Partially correct but has logical gaps
     or is not well-supported.
- 4: Mostly logical and relevant, with minor
     flaws.
- 5: Perfectly logical, coherent, and directly
     justifies the labels based on the text.

Your answer MUST be a single digit from 1 to 5.

Original Text:
---
{text}
---

Predicted Labels: {labels}
---

Reasoning:
---
{reasoning}
---

Based on your evaluation, what is the score
for this reasoning? Answer with a single digit
(1-5).
Score:"""
\end{lstlisting}

\section{Prompt for Self-Consistency (SC)}
\label{app:sc_prompt}

\begin{lstlisting}[style=promptcompact, caption={Prompt for self-consistency (SC).}, label={lst:sc-prompt}]
prompt_template = """
You are an expert in patent classification
with deep knowledge of the Cooperative Patent
Classification (CPC) system. Your task is to
analyze the provided patent text (abstract or
representative claim) and identify the relevant
CPC technical subclasses. You must also provide
a detailed, logical reasoning process explaining
why each subclass was selected, ensuring
alignment with the CPC class definitions.

Instructions:
1. Identify one or more CPC subclasses (e.g.,
   'G06F', 'H04L', 'A61B') that best represent
   the technical content of the patent text.
   Ensure the subclasses are valid and specific
   to at least the 4-character CPC code level.
2. Provide a rigorous reasoning process that
   justifies your classification decisions.
   The reasoning must be less than 60 words and
   must be coherent, grounded in the technical
   content of the patent, and aligned with the
   CPC subclass definitions.
3. Consider that a patent may belong to multiple
   CPC subclasses due to its multi-label nature.
4. Output your response strictly in the JSON
   format below, with no additional text,
   comments, or Markdown formatting.

Example output format:
{
  "predicted_labels": ["B25B"],
  "reasoning": "The patent text describes a
    mechanical hand tool, specifically a wrench.
    Key features include 'adjustable jaws' for
    gripping 'nuts and bolts' and a 'handle' for
    applying torque, aligning with CPC subclass
    B25B."
}

Patent text:
---
{patent_text}
"""
\end{lstlisting}

\end{document}